\documentclass[journal]{IEEEtai}

\usepackage[colorlinks,urlcolor=blue,linkcolor=blue,citecolor=blue]{hyperref}

\usepackage{textcomp}
\usepackage{url}
\usepackage{verbatim}
\usepackage{graphicx}
\usepackage{cite}
\usepackage{algorithmic}
\usepackage{algorithm}
\usepackage{array}
\usepackage{textcomp}
\usepackage{url}
\usepackage{verbatim}
\usepackage{graphicx}
\usepackage{cite}
\usepackage{amsmath,amssymb,amsfonts}
\usepackage{multirow}
\usepackage{graphicx}
\usepackage{lipsum}
\usepackage{float}
\usepackage{subcaption} 
\usepackage{caption} 
\usepackage{textcomp}
\usepackage{booktabs}

\begin{document}

\title{An Uncertainty-Aware Loss Function Incorporating Fuzzy Logic: Application to MRI Brain Image Segmentation}

\author{Hanuman Verma, Akshansh Gupta, Pranabesh Maji, Saurav Mandal, Vijay Kumar Pandey
	\thanks{This paragraph of the first footnote will contain the date on which
		you submitted your paper for review. It will also contain support information,
		including sponsor and financial support acknowledgment. For example, 
		``This work was supported in part by the U.S. Department of Commerce under Grant BS123456.'' }
	\thanks{Hanuman Verma is with the Department of Mathematics, Bareilly College, Bareilly (MJP Rohilkhand University), Uttar Pradesh, 243005, India (e-mail: hv4231@gmail.com). }
	\thanks{Akshansh Gupta is with the CSIR-Central Electronics Engineering Research Institute, Pilani, Rajasthan, 333031, India (e-mail: akshanshgupta83@gmail.com).}
	\thanks{Pranabesh Maji is with the CSIR-Central Electronics Engineering Research Institute, Pilani, Rajasthan, 333031, India (e-mail:pranabesh@ceeri.res.in).}
	\thanks{Saurav Mandal is with the CICMR, Regional Medical Research Center, Dibrugarh, India (e-mail: saurav.mandal@icmr.gov.in).}
	\thanks{Vijay Kumar Pandey is with the Department of Statistics, Bareilly College, Bareilly, Uttar Pradesh, 243005, India (e-mail: vijay.pandey550@gmail.com).}}


\maketitle

	\begin{abstract}
	Accurate brain image segmentation, particularly for distinguishing various tissues from magnetic resonance imaging (MRI) images, plays a pivotal role in finding the neurological disease and medical image computing. In deep learning approaches, loss functions are very crucial for optimizing the model. In this study, we introduce a novel loss function integrating fuzzy logic to deals uncertainty issues in brain image segmentation into various tissues. It integrates the well-known categorical cross-entropy (CCE) loss function and fuzzy entropy based on fuzzy logic. By employing fuzzy logic, this loss function accounts for the inherent uncertainties in pixel classifications. The proposed loss function has been evaluated on two publicly available benchmark datasets, IBSR and OASIS, using two widely recognised architectures, U-Net and U-Net++. Experimental results demonstrate that the trained model with proposed loss function provided better results in comparison to the CCE optimisation function in terms of various performance metrics. Additionally, it effectively enhances segmentation performance while handling meaningful uncertainty during training. The findings suggest that this approach not only improves segmentation outcomes but also contributes to the reliability of model predictions.
\end{abstract}


\section{Introduction}
\IEEEPARstart{B}{rain} image segmentation from magnetic resonance imaging (MRI) images in various tissues is a critical task in medical image computing and has significant implications for patient diagnosis, treatment planning, and monitoring. Manual segmentation by domain experts is laborious and subject to inter-observer variability. As a result, there is a growing demand for accurate and automated segmentation algorithms to assist clinicians in this process. Many image segmentation algorithms based on unsupervised and supervised learning are used to segment the MRI brain image \cite{Verma2016,Fawzi2021,Gupta2023,Wang2022}. In recent, one of the most promising approaches in this domain is the use of deep learning is convolutional neural networks (CNNs) architecture, particularly the U-Net architecture, which has demonstrated state-of-the-art performance in a variety of medical image segmentation tasks \cite{Azad2024,Punn2022,Liu2020}. The original U-Net architecture, proposed by Ronneberger et al. \cite{Ronneberger2015}, is a CNN that combines a contracting path to capture context and an expansive path to enable precise localization. Moreover, the advancements that have been made most recently in the realm of deep learning have led to the development of modified iterations of the U-Net architecture. While the standard U-Net architecture has been widely adopted, recent studies have explored various modifications and enhancements to improve its performance. 

Fuzzy logic referred from the fuzzy set theory deals with reasoning that is approximate rather than fixed and exact in classical set theory. Unlike classical set theory, where data can only take binary values that is, true or false, whereas fuzzy logic allows data to have a range of values between 0 and 1, representing degrees of membership (belongingness) \cite{Zadeh1965,Chaira2015}. This makes it especially useful in situations where data is imprecise. It is used in image segmentation, especially when dealing with the inherent uncertainty and imprecision in medical imaging data. A well-known unsupervised fuzzy clustering technique based on fuzzy logic, named fuzzy c-means algorithm \cite{Bezdek1981},
are commonly used for image segmentation \cite{Verma2014, Hu2021, Verma2023}. These methods assign degrees of membership to each pixel, allowing for soft classification rather than hard segmentation. Intuitionistic fuzzy c-means based on intuitionistic fuzzy set theory is an advanced variant of FCM that incorporates a hesitation to better handle ambiguity in pixel classification. It introduces an additional form of uncertainty known as hesitation and enhances segmentation accuracy, especially in complex regions of brain images \cite{Verma2016, Verma2020, Yang2021}. 

The CNNs architecture and its variants is based on crisp classifications and the recent developments in the integration of fuzzy logic \cite{Das2020} with CNN have been addressed in image segmentation particularly in handling uncertainty. U-Net architecture, variant of CNN is used for image segmentation and provides the promising results. However, U-Net is unable to handle uncertainty caused by vagueness and the partial volume effect. Some efforts have been made in the literature to integrate fuzzy logic into deep learning to address this uncertainty. The integration of a fuzzy layer in deep learning was introduced by Price et al., \cite{Price2019} using the Choquet and Sugeno fuzzy integrals. This approach exploits the aggregation properties in deep learning and is applied to semantic segmentation. Sharma et al. \cite{Sharma2019} introduced a novel pooling layer based on fuzzy logic within the convolutional layer to extract useful information. Huang et al.\cite{Huang2021} introduced the concept of fuzzy logic in convolutional neural networks for the segmentation of ultrasound breast images. First, the image is augmented using wavelet transform, and then the augmented image was transformed into the fuzzy domain with various membership functions to handle the uncertainty in the images. Badawy et al. \cite{Badawy2021} proposed a deep learning model that combines fuzzy logic with a convolutional neural network for the semantic segmentation of breast tumors from ultrasound images. In this approach, the preprocessing steps are carried out using fuzzy logic, while the segmentation is performed using the CNN. Huang et al. \cite{Huang2022} proposed a method for breast image segmentation from ultrasound images using fuzzification to reduce uncertainty in convolutional neural networks. This approach, named the spatial and channel-wise fuzzy uncertainty reduction network, aims to minimize uncertainty in image segmentation. Subhashini et al. \cite{Subhashini2022} integrates the concepts the fuzzy logic and deep learning model together to make a three-way decision to solve the uncertainty problem arises due to vagueness. To incorporate fuzzy logic into the U-Net architecture, Chen et al.\cite{Chen2022} designed the target-aware U-Net model with fuzzy skip connections and applied it to pancreas segmentation. The fuzzy skip connection employs fuzzy logic to extract high-level semantic features. Additionally, a target mechanism is incorporated into the decoder part of this model to enhance feature representation for the desired segmentation. Karaköse (2024), \cite{Karakose2024}, suggested a method based on fuzzy cognitive map integration with deep learning models using improved loss function and used for images classification, where a custom loss function has been defined with a special combination of loss functions, and the results have been evaluated. 

The loss function (minimization of objective function) measures the difference between the predicted output of a training model and actual value of the training data. It computes the error of the model and guides the optimization process by providing error through loss function during training and objective is to minimizing the loss function which ensures the prediction of model become more accurate over time. It is also critical components of backpropagation algorithm, which is used to compute the gradient of the loss with respect to model parameters.  The various loss function introduced to improve the model performance  \cite{Azad2023,Jadon2020,Janocha2017}. Among these the categorical cross-entropy (CCE) is a loss function commonly used in classification tasks where the goal is to classify input data into one of multiple classes. It computes the difference between the predicted probability distribution of output of the model and the true distribution that is the actual labels. An uncertainty-aware cross-entropy loss for the semantic segmentation that integrate the dynamic uncertainty estimates has been proposed by Landgraf et al.\cite{Landgraf2023}. It incorporates predictive uncertainties pixelwise weighting of the regular cross-entropy loss function, based on Monte Carlo Dropout during the training process.

In this work, we propose a novel fuzzy categorical cross entropy (FCCE) loss function designed to address data uncertainty and enhance segmentation performance. The main contribution in this work includes:

\begin{itemize}
	\item It integrates the widely-used categorical cross entropy with fuzzy entropy principles derived from fuzzy logic. 
	\item The degree of membership, essential for incorporating fuzziness, is calculated using the fuzzy c-means algorithm. 
	\item With leveraging fuzzy entropy, the proposed method effectively captures and utilizes uncertainty and improve model robustness and segmentation accuracy.
	\item It enhances segmentation accuracy by effectively handling the uncertainties.
	
\end{itemize}

To evaluate the effectiveness of the FCCE loss function, experiments are conducted on two publicly available MRI brain datasets: the Internet Brain Segmentation Repository (IBSR) and the Open Access Series of Imaging Studies (OASIS). Segmentation performance was assessed using multiple metrics, including accuracy, Dice coefficient, and Intersection over Union (IoU), as well as their corresponding validation measures such as validation accuracy, validation Dice coefficient, and validation IoU. These metrics comprehensively demonstrate the capability of the FCCE approach in delivering improved segmentation outcomes.

The rest of this paper is organized as follows: Section \ref{Pre} presents the preliminaries and foundational concepts. In Section \ref{Pro}, we detail the proposed fuzzy categorical cross entropy loss function. The structure of U-Net and U-Net ++ is described in the Section \ref{Me}. Section \ref{Exp} describes the experimental setup, followed by Section \ref{RD}, which presents the results and their analysis. Finally, Section \ref{con} concludes the work and discusses future directions.

\section{Preliminaries}\label{Pre}
\subsection{Fuzzy logic and fuzzy c-means algorithm}
A fuzzy set A on the set X with membership degree $\mu_A : X \to [0,1]$ is defined as \cite{Zadeh1965}: $A = \{ \langle x, \mu_A(x) \rangle : \forall x \in X \}$. The membership degree represents the degree of belongingness with the available information in imprecise data. It handles the uncertainty in uncertain data. In fuzzy set, non-membership degree (non-belongingness) is computed as one minus the membership degree. 

A well-known clustering algorithm based on fuzzy logic is fuzzy c-means (FCM) algorithm introduced by \cite{Bezdek1981}. It is used in many applications such as image segmentation and clustering and useful in ambiguous data. FCM is a non-linear optimization problem with objective function comprises the Euclidean distance between centroids and pixels in the image, and comprises membership degree with fuzzy fuzzifier constant. The goal is to minimize the objective function to minimize the error and find the promising results. Mathematically, it is written as \cite{Bezdek1981}:  

\begin{equation}
	\min J_{\text{FCM}}(U, V : X) = \sum_{i=1}^{c} \sum_{j=1}^{N} u_{ij}^m d^2(x_j, v_i)
	\label{eq:1}
\end{equation}

where \( X = \{x_1, x_2, \dots, x_N\} \) is the image containing \( N \) pixels, \( c \) (\( 1 < c < N \)) is the number of clusters, \( m \) (\( m > 1 \)) is the fuzzifier constant, \( V = (v_1, v_2, \dots, v_c) \) is the centroid having center of \( c \) clusters, and \( d^2(x_j, v_i) = (x_j - v_i)^2 \) is the Euclidean distance in two dimensions between the pixels \( x_j \) and centroid \( v_i \). \( u_{ij} \subseteq U_{(c \times N)} \) is the membership degree of the pixel points \( x_j \) to the \( i \)-th cluster center \( v_i \), with \( \sum_{i=1}^{c} u_{ij} = 1 \) and \( u_{ij} \in [0, 1] \). The FCM objective function \eqref{eq:1} , with \( \sum_{i=1}^{c} u_{ij} = 1 \) and each \( u_{ij} \in [0, 1] \), can be minimized with the following extrema \cite{Bezdek1981}:

\begin{equation}
	u_{ij} = \frac{1}{\left( \sum_{r=1}^{c} \left( \frac{d(x_j, v_i)}{d(x_j, v_r)} \right)^{\frac{2}{m-1}} \right)}, \quad \forall i, j
	\label{eq:2}	
\end{equation}

\begin{equation}
	v_i = \frac{\sum_{j=1}^{N} u_{ij}^m x_j}{\sum_{j=1}^{N} u_{ij}^m}, \quad \forall i	
	\label{eq:3}
\end{equation} 

It is iterative algorithm with updating the centroids and membership degree and improve the degree of belongingness. 
\subsection{Categorial cross-entropy loss function}
The categorical cross-entropy is a widely used as a loss function in multi-class classification problems \cite{Azad2023}. In the context of image classification, it computes how well the predicted class probabilities match the true labels of the class. It computes the difference between two probability distributions that is the true label distribution $y_i$ (one-hot encoded) and the predicted label distribution $\left( \hat{y}_i \right)$, obtained from the model training. It is computed as:

\begin{equation}
	L_{\text{CCE}} = -\sum_{i=1}^c y_i \log(\hat{y_i})
	\label{eq:4}	
\end{equation} 
where $c$ is the number of classes. The partial derivative of loss function to the logits \( (z_i) \) is obtained from the softmax activation function. The softmax activation function compute the $z_i$ to the probabilities $\hat{y}_i = \frac{e^{z_i}}{\sum_{i=1}^c z_i}$ . Using the chain rule, gradient of \( L_{\text{CCE}} \) is computed as:

\begin{equation}
	\frac{\partial L_{\text{CCE}}}{\partial z_i} = \frac{\partial L_{\text{CCE}}}{\partial \hat{y}_i} \cdot \frac{\partial \hat{y}_i}{\partial z_i}	
	\label{eq:5}
\end{equation} 

\section{Proposed novel loss function}\label{Pro}
\textbf{Combining categorical cross entropy with fuzzy entropy:}

The loss function measures the model performance that how well the designed model’s prediction outcome and reference outcomes are matches. The objective is to minimize the loss function (optimizing the objective) to improve the performance of model and neural network learn to make prediction that are close as possible to the ground truth data. It improves the learning process with assigning the suitable value to the loss. The categorical cross-entropy work on basis of classical set theory and unable to handle the uncertainty during the training process. To overcome these issues, we present fuzzy categorical cross entropy with fuzzy entropy based on fuzzy logic is to handle the uncertainty in vague kind of situation. The uncertainty measure is a new viewpoint for decision making from the ambiguous information. In medical images inherently possess uncertainty due to factors such as noise, complex nature, partial volume effect, and inconsistency. The new loss function is defined in following subsections:

\subsection{Fuzzy entropy function}
The fuzzy entropy derived from the Shannon entropy \cite{Shannon1948} in the context of fuzzy set. It measures the uncertainty or fuzziness in a fuzzy set with the degree of belongingness (membership degree). The Shannon entropy in term of fuzzy set is given by: 

\begin{equation}
	L_{\text{FE}} = -\sum_{i=1}^c \sum_{j=1}^N u_{ij} \log(u_{ij})	
	\label{eq:6}
\end{equation} 
where \( u_{ij} \) is the membership degree for the $ith$ class to $jth$ data and computed using the FCM algorithm through the equation \eqref{eq:2} . It sums the uncertainty across all data in the fuzzy sets. The higher value of fuzzy entropy shows the set is more uncertain whereas the less value of entropy indicates the less uncertain. Using the $log$ value offers a less penalty for the small differences between the actual and predicted value. 

In machine learning, the fuzzy entropy function to compute the loss function during the model training. The objective is to minimize the uncertainty between the data through the fuzzy entropy function, and considered as a loss function to optimizes the model performance. The gradient which analyses how the entropy changes with respect to changes of the membership degree and used in optimization. The gradient of the loss function with respect to weight $(w)$ of the model during backpropagation. It involves the partial derivative to compute the gradient using the chain rule:

\begin{equation}
	\frac{\partial L_{\text{FE}}}{\partial w} = -\sum_{i=1}^c \sum_{j=1}^N \frac{\partial L_{\text{FE}}}{\partial u_{ij}} \cdot \frac{\partial u_{ij}}{\partial c_j}
	\label{eq:7}
\end{equation} 

With computing the gradient, the weight can be updated using the optimization algorithm like gradient decent. The gradient of the fuzzy entropy is  $\frac{\partial L_{\text{FE}}}{\partial u_{ij}} = -\log(u_{ij}) - 1$ and $\frac{\partial u_{ij}}{\partial c_j}$ is computed using the \eqref{eq:2} and \eqref{eq:3}.
\subsection{Fuzzy categorical cross entropy}

The fuzzy categorical cross entropy $(L_{ FCCE} )$  loss functions is the combination of categorical cross-entropy  $(L_{ CCE} )$ and fuzzy entropy $(L_{FE} )$, which integrate the fuzzy logic and is mathematically defined as:

\begin{equation}
	L_{\text{FCCE}} = -\sum_{i=1}^c y_i \log(\hat{y_i}) -\sum_{i=1}^c \sum_{j=1}^N u_{ij} \log(u_{ij})
	\label{eq:8}
\end{equation} 

\begin{figure}[H]
	\centering
	\includegraphics[width=0.5\textwidth]{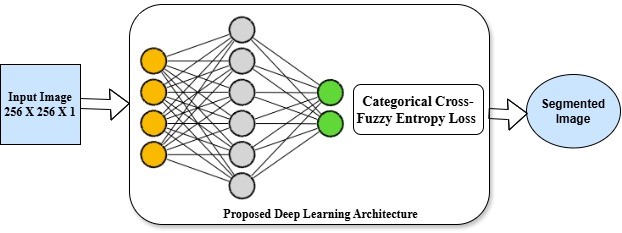}
	\caption{Deep learning architecture with proposed fuzzy categorial cross entropy for brain image segmentation. }
	\label{fig:fign1}
\end{figure}

During the optimizing the model, it computed the error between the predicted image and manual segmented image and also find the fuzzy entropy using FCM algorithm with degree of belongingness, how much that is to belong the similar class and non-belonging to dissimilar class. With $L_{\text{FE}}$ , it computes the entropy using unsupervised FCM clustering algorithm along with unlabelled data and $L_{\text{CCE}}$ along with the labelled data. It treated the loss using labelled and unlabelled data as supervised and unsupervised learning ways. The gradient of fuzzy categorical cross entropy loss functions that is $ \frac{\partial L_{\text{FCCE}}}{\partial w}$ is computed with combine the gradient of categorical cross-entropy $(L_{\text{CCE}})$  and fuzzy entropy $(L_{\text{FE}})$. The fuzzy categorical cross entropy loss functions are incorporate in the deep learning structure shown in Fig. \ref{fig:fign1}.

\section{Methodology}\label{Me}
The essential concepts, definitions, and notation are explained in the following subsection:

\subsection{U-Net architecture }
\begin{figure}[H]
	\centering
	\includegraphics[width=0.5\textwidth]{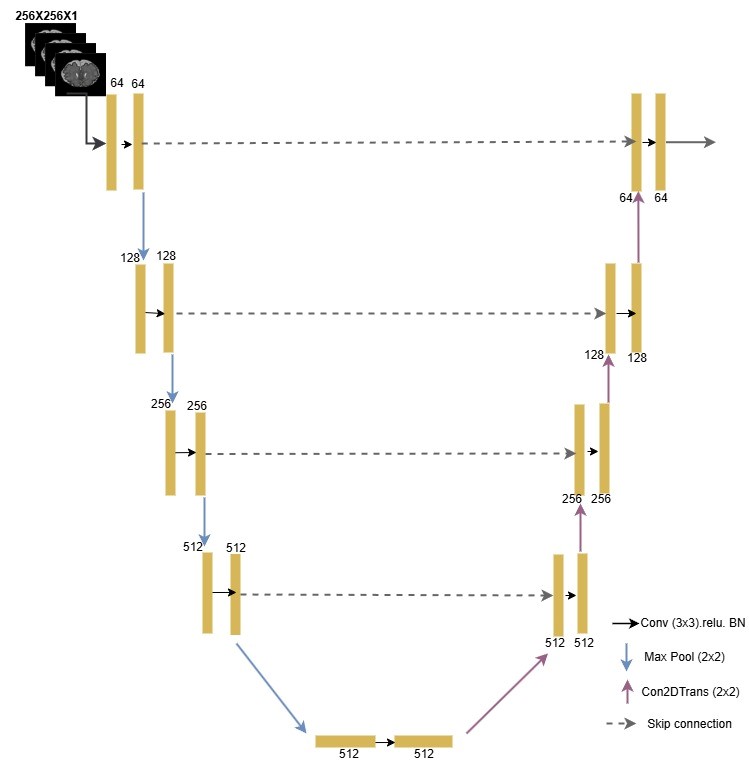}
	\caption{U-Net architecture featuring various neurons and 2D convolution with a kernel size of $3\times 3$, followed by max pooling with a kernel size of $2\times 2 $. The ReLU activation function is used, followed by Batch normalization.}
	\label{fig:fign2}
\end{figure}

Figure \ref{fig:fign2}, illustrates the U-Net network architecture \cite{Ronneberger2015}. The framework comprises a contracting pathway on the left side and an expanding pathway on the right side. The contracting path adheres to the conventional structure of a convolutional network. The process involves iteratively applying two $3\times 3$ convolutions (without padding), each followed by a rectified linear unit (ReLU) and a $2\times 2$ max pooling operation with a stride of $2$ to down sample the data. During each down sampling step, we increase the number of feature channels by a factor of two. Each step in the expansive path involves up sampling the feature map and then applying a $2\times 2$ convolution ("up-convolution") to reduce the number of feature channels by half. This is followed by concatenating the resulting feature map with the corresponding cropped feature map from the contracting path. Finally, two $3\times 3$  convolutions are applied, each followed by a rectified linear unit (ReLU). Cropping is essential since border pixels are lost during each convolution. A $1\times 1$ convolution is employed at the last layer to transform each $64$-component feature vector into the required number of classes. The network consists of a total of $23$ convolutional layers \cite{Ronneberger2015}.

\subsection{U-Net++  architecture }
\begin{figure}[h!]
	\centering
	\includegraphics[width=0.5\textwidth]{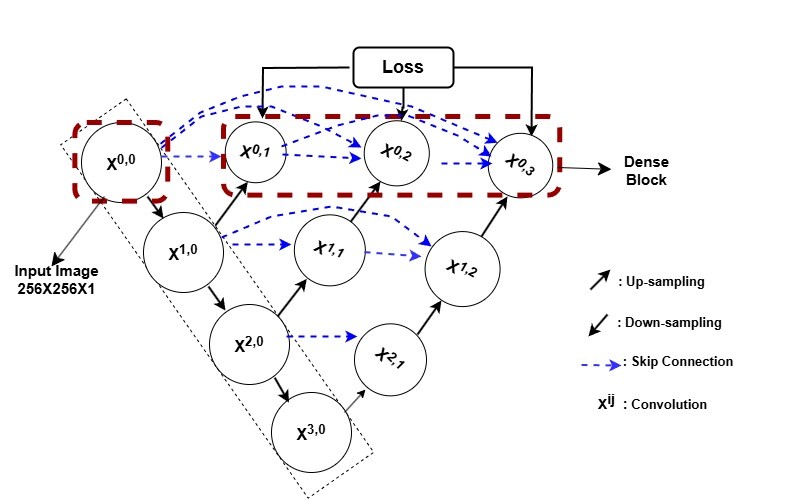}
	\caption{U-Net++ architecture with the skip connectionsn.}
	\label{fig:fign3}
\end{figure}

The authors Zhou et al. \cite{Zhou2018, Zhou2019} present U-Net++, a novel segmentation architecture that makes use of layered and dense skip connections, in order to improve the accuracy of image segmentation in medical pictures and capture complex properties of target objects. This architecture's basic premise is that by gradually improving the high-resolution feature maps from the encoder network before combining them with the semantically rich feature maps from the decoder network, the model will be better able to capture the finer details of the foreground objects. The authors claim that if the feature maps produced by the encoder and decoder networks show semantic similarity, the learning task of the network would be made easier.  In contrast to the simple skip connections typically employed in U-Net, which directly pass high-resolution feature maps from the encoder to the decoder network, resulting in the merging of feature maps with different meanings. U-Net++ is structured with an initial encoder sub-network or backbone, which is then followed by a decoder sub-network. The key differentiating factor between U-Net++ and U-Net is the incorporation of re-designed skip paths, depicted in green and blue, which connect the two sub-networks. Additionally, U-Net++ utilizes deep supervision, represented by the colour red, shown in Fig. \ref{fig:fign3}.

In the concise way, there are three ways that U-Net++ is different from the original U-Net. The first is having convolution layers on skip pathways (shown in green), which connect the encoder and decoder feature maps semantically. The second is having thick skip connections on skip pathways (shown in blue), which makes gradient flow better. And the third is having deep supervision (shown in red).
\begin{equation}
	x^{i,j} = 
	\begin{cases}
		H(D(x^{i-1,j})), & j = 0 \\
		H\left( \left[ x^{i,k} \right]_{k=0}^{j-1}, U(x^{i+1,j-1}) \right), & j > 0
	\end{cases}
	\label{eq:9}
\end{equation} 

\begin{equation}
	L(Y, P) = -\frac{1}{N} \sum_{c=1}^C \sum_{n=1}^N \left( y_{n,c} \log p_{n,c} + \frac{2y_{n,c} p_{n,c}}{y_{n,c}^2 + p_{n,c}^2} \right)
	\label{eq:10}
\end{equation} 

\section{Experiment }\label{Exp}
\subsection{Dataset }: In this study, the public MRI brain datasets, namely IBSR (Internet Brain Segmentation Repository) and OASIS (Open Access Series of Imaging Studies), serve as pivotal resources in neuroimaging research, offering unique and complementary data. IBSR provides high-resolution MRI scans with detailed ground truth annotations, enabling precise evaluation of segmentation algorithms and insights into brain anatomy. In contrast, OASIS includes a diverse collection of MRI scans from healthy individuals and patients with cognitive impairments, along with rich clinical metadata, facilitating the study of brain pathology, age-related changes, and disease progression. Together, these datasets provide a comprehensive foundation for advancing segmentation techniques and understanding neurological conditions. \\
\textbf{IBSR}: Internet brain segmentation repository (https://www.nitrc.org/projects/ibsr/) is used to evaluate the performance of new segmentation technique, as it provides the manually guided expert segmentation brain data. IBSR consist of 20 normal T1-Weighted real MRI brain data along with corresponding manual segmentation brain data. The brain dataset is well-annotated with segmentation labels that detail various brain structures and tissues. These labels are essential for training and validating segmentation algorithms, as they provide a ground truth against which the performance of automated methods can be assessed. It is provided by the Center for Morphometric Analysis at Massachusetts General Hospital. \\
\textbf{OASIS}: Open access series of imaging studies (http://www.oasis-brains.org/ ) provides brain imaging data which is publicly available for distribution and data analysis. It comprises the 416 subject aged 18 to 96 and each subject 3 or 4 individual T1-weighted MRI scans obtained within a single imaging session are included. The non-brain tissues are removed using the using the Brain Extraction Tool. The brain dataset is well-annotated with various segmentation labels that detail various brain structures and brain tissues.
\subsection{Baseline and Implementation }
To implement the fuzzy categorical cross entropy $(L_{{FCCE}})$  loss optimization function, we compute the membership degree through FCM algorithm and further compute the fuzzy entropy $(L_{{FE}})$. The categorical cross-entropy $(L_{{CCE}})$ computed using the standard function \eqref{eq:4} in deep learning. For comparison of proposed  $L_{\text{FCCE}}$ optimization function with $L_{{CCE}}$, we use U-Net \cite{Ronneberger2015} and U-Net++ architecture \citen{Zhou2019} for MRI brain image segmentation task. We choose the fuzzy entropy in $L_{{FCCE}}$ using fuzzy logic and check the performance of fuzzy categorical cross entropy, to achieve the promising accuracy, as depicted in Fig. \ref{fig:fign1}. We use the same structure of U-Net and U-Net++ for  $L_{{CCE}}$ and $L_{{FCCE}}$ optimization function. Relu activation function and Adam optimizer is used in both U-Net and U-Net++ along with Batch normalization and Dropout layers after each convolution layer in U-Net++ architecture to overcome the overfitting. And batch normalization layers are integrated with the U-Net architecture. The kernel size  $3\times 3$ is used in the both U-Net and U-Net++ and details of neuron are shown in Fig. \ref{fig:fign2} and Fig. \ref{fig:fign3}.  The experiments are implemented in Keras with Tensorflow backend. To evaluate the experimental results, the measurement metrics such as accuracy $(AC)$, Dice coefficient $(DC)$ and Intersection over Union $(IoU)$ along with the validation results like accuracy validation $({AC}_{{val}})$, Dice coefficient validation $({DC}_{{val}})$ and Intersection over Union validation $({IoU}_{{val}})$ are used to check the performance of the loss function.   

\section{Results and discussion }\label{RD}

\begin{figure*}  
	\centering
	\begin{subfigure}[t]{0.45\textwidth}
		\centering
		\includegraphics[width=\textwidth]{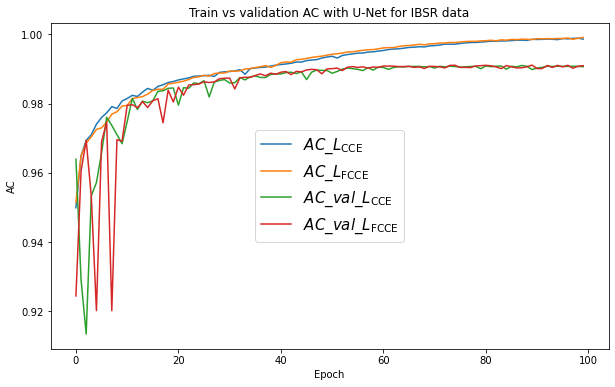}
	\end{subfigure}
	\hfill
	\begin{subfigure}[t]{0.45\textwidth}
		\centering
		\includegraphics[width=\textwidth]{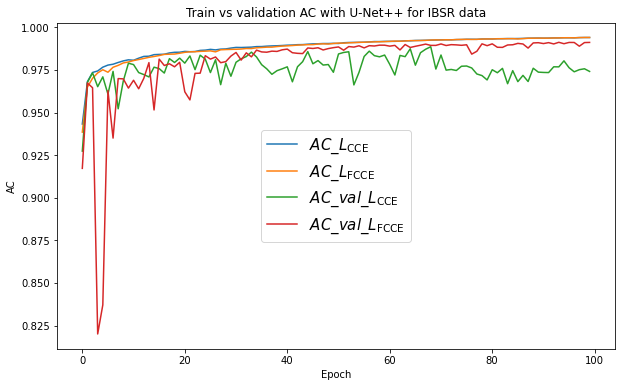}
	\end{subfigure}
	
	\vspace{2em}
	
	\begin{subfigure}[t]{0.45\textwidth}
		\centering
		\includegraphics[width=\textwidth]{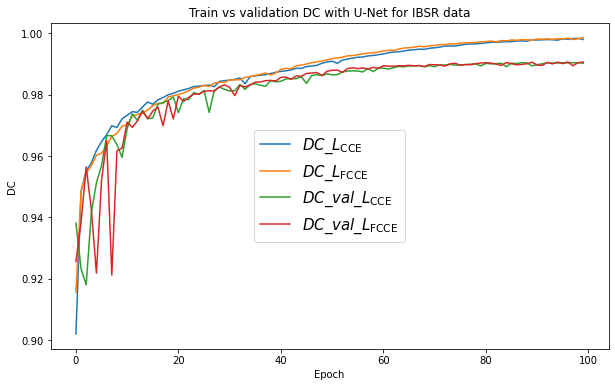}
	\end{subfigure}
	\hfill
	\begin{subfigure}[t]{0.45\textwidth}
		\centering
		\includegraphics[width=\textwidth]{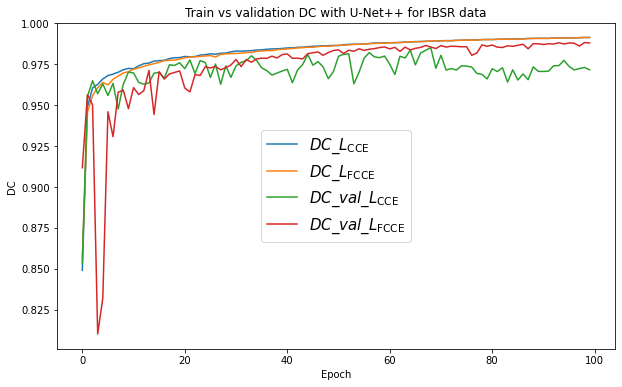}
	\end{subfigure}
	
	\vspace{2em}
	
	\begin{subfigure}[t]{0.45\textwidth}
		\centering
		\includegraphics[width=\textwidth]{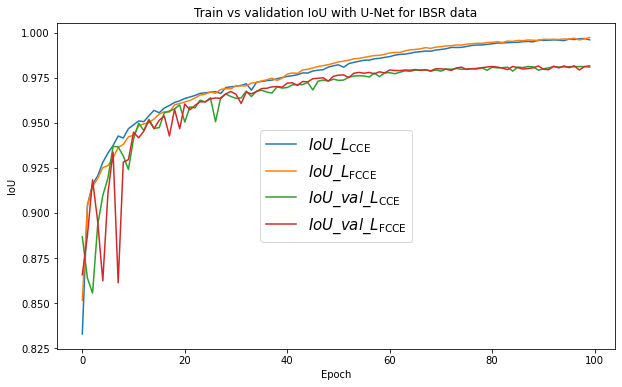}
	\end{subfigure}
	\hfill
	\begin{subfigure}[t]{0.45\textwidth}
		\centering
		\includegraphics[width=\textwidth]{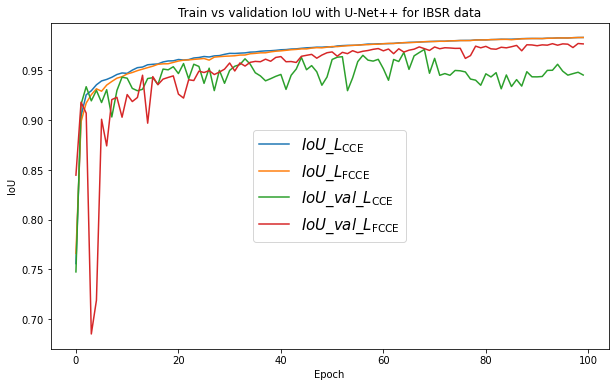}
	\end{subfigure}
	
	\caption{Plot for training versus validation performance for U-Net and U-Net++ architecture for IBSR data with categorical cross-entropy $(L_{CCE})$  and  fuzzy categorical cross entropy $(L_{FCCE})$  loss optimization function along with epoch in term of performance metrices: accuracy, Dice coefficient and Intersection over Union. The performance of U-Net and U-Net++ with $L_{FCCE}$ in term of $AC$,$DC$,$IoU$ performance slightly better with the epoch. During training the model, the U-Net validation with $L_{FCCE}$ in term of  ${AC}_{{val}}$, ${DC}_{{val}}$, ${IoU}_{{val}}$  providing better results with optimizing with fuzzy logic. The U-Net++ validation with $L_{FCCE}$ in term of ${AC}_{{val}}$, ${DC}_{{val}}$, ${IoU}_{{val}}$ providing more better results in comparison to $L_{CCE}$. The fuzzy categorical cross entropy $(L_{FCCE})$ optimizing the objective with the quantifying the uncertainty during training the model.}
	\label{fig:fign4}
\end{figure*}

Ablation study with categorical cross-entropy $\left( L_{CCE} \right)$  and  fuzzy categorical cross entropy $\left( L_{FCCE} \right)$  loss optimization function with fuzzy logic presented in this study on IBSR and OASIS brain dataset. Firstly, we test the U-Net and U-Net++ architecture with $ L_{CCE}$  and $L_{FCCE}$ for segmentation on IBSR data. The 05 cases from the IBSR dataset have been taken and from these cases $1280$ brain 2-D images is obtained for the training the model. The corresponding labelled image is used for the validation the model. Table \ref{tab:tabn1}, comprises the segmentation results and compare the efficacy of segmentation performance in term of accuracy $(AC)$, Dice coefficient $(DC)$ and Intersection over Union $(IoU)$ and accuracy validation $({AC}_{{val}})$, Dice coefficient validation $({DC}_{{val}})$, Intersection over Union validation $({IoU}_{{val}})$. As seen, the U-Net and U-Net++ architecture performed better with $L_{FCCE}$ loss function in comparison to$L_{CCE}$ loss function. The $L_{FCCE}$ integrates the fuzzy logic with belongingness with predicted value during training the model and able to handle the kind of uncertainty. The U-Net++ promising the better validation results with fuzzy categorical cross entropy during the model training. The fuzzy set theory handles the boundary vagueness and other kind of information. The training and validation performance during training the model with $L_{CCE}$  and $L_{FCCE}$ is depicted in Fig. \ref{fig:fign4} with the various parameters for IBSR brain data.

\begin{table}[H]
	\centering
	\caption{U-Net and U-Net++ architecture segmentation performance with $L_{CCE}$ and $L_{FCCE}$ loss function for IBSR data. We take the number of epochs 100 and Batch size 2.  And use the segmentation metrices as accuracy $(AC)$, Dice coefficient $(DC)$ and Intersection over Union $(IoU)$ and their corresponding validations.}
	\label{tab:tabn1}
	\begin{tabular}{llcc}
		\toprule
		{Algorithm} & {Metrics} & {$L_{CCE}$} & {$L_{FCCE}$} \\
		\midrule
		U-Net    & AC             & 0.9986 & 0.9991 \\
		& $AC_{val}$ & 0.9907 & 0.9910 \\
		& DC             & 0.9980 & 0.9986 \\
		& $DC_{val}$ & 0.9903 & 0.9906 \\
		& IoU            & 0.9960 & 0.9972 \\
		& $IoU_{val}$ & 0.9808 & 0.9815 \\
		U-Net++  & AC             & 0.9940 & 0.9941 \\
		& $AC_{val}$ & 0.9741 & 0.9911 \\
		& DC             & 0.9914 & 0.9915 \\
		& $DC_{val}$ & 0.9717 & 0.9881 \\
		& IoU            & 0.9831 & 0.9832 \\
		& $IoU_{val}$ & 0.9454 & 0.9767 \\
		\bottomrule
	\end{tabular}
\end{table}

\begin{figure}[H]
	\centering
	\includegraphics[width=0.45\textwidth]{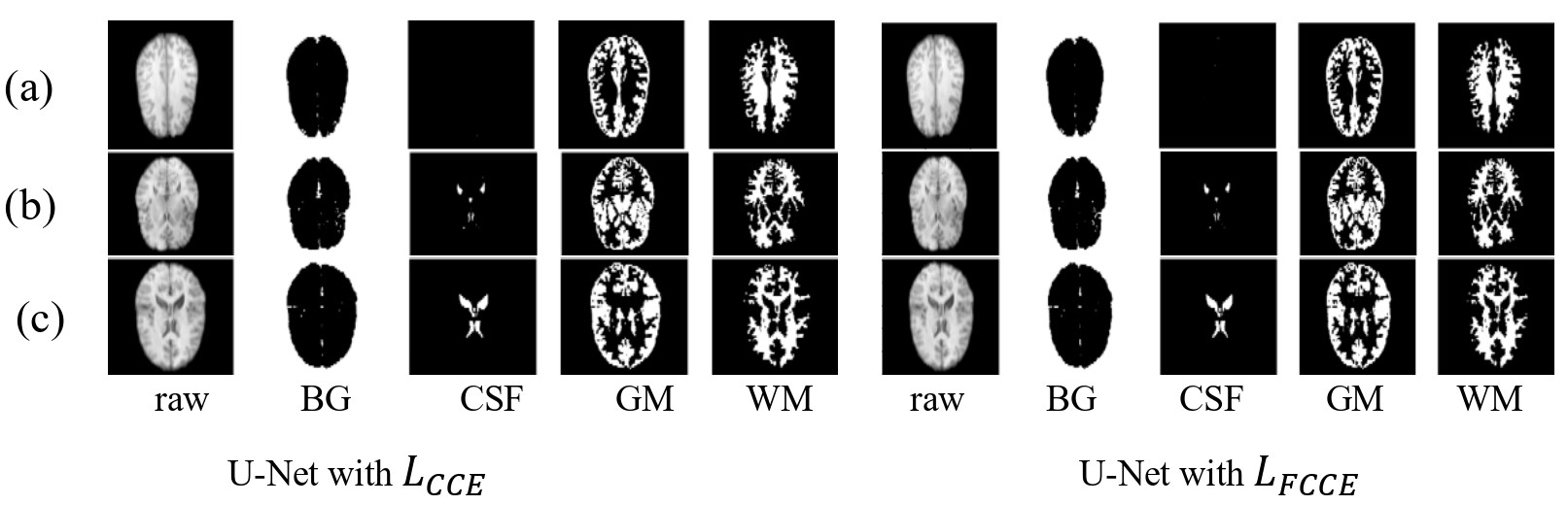}
	\caption{Qualitative segmentation results using U-Net with  $L_{CCE}$, and U-Net with  $L_{FCCE}$ (proposed loss function) on the IBSR brain data for the various image (a), (b) and (c). The most left row, comprised the raw image and corresponding predicted image comprises with various tissues such as background (BG), CSF, GM, and WM in this study. U-Net architecture with loss function is trained and designed and then predicted and depicted for same images (a), (b) and (c) respectively. The U-Net with  $L_{FCCE}$ shows the better segmentation in CSF, GM and WM tissues as it handles the uncertainty at the boundary region, as shown in case of CSF tissues. The boundary pixels in CSF tissues are changing which can be promising with uncertainty. }
	\label{fig:fign5}
\end{figure}

\begin{figure}[H]
	\centering
	\includegraphics[width=0.45\textwidth]{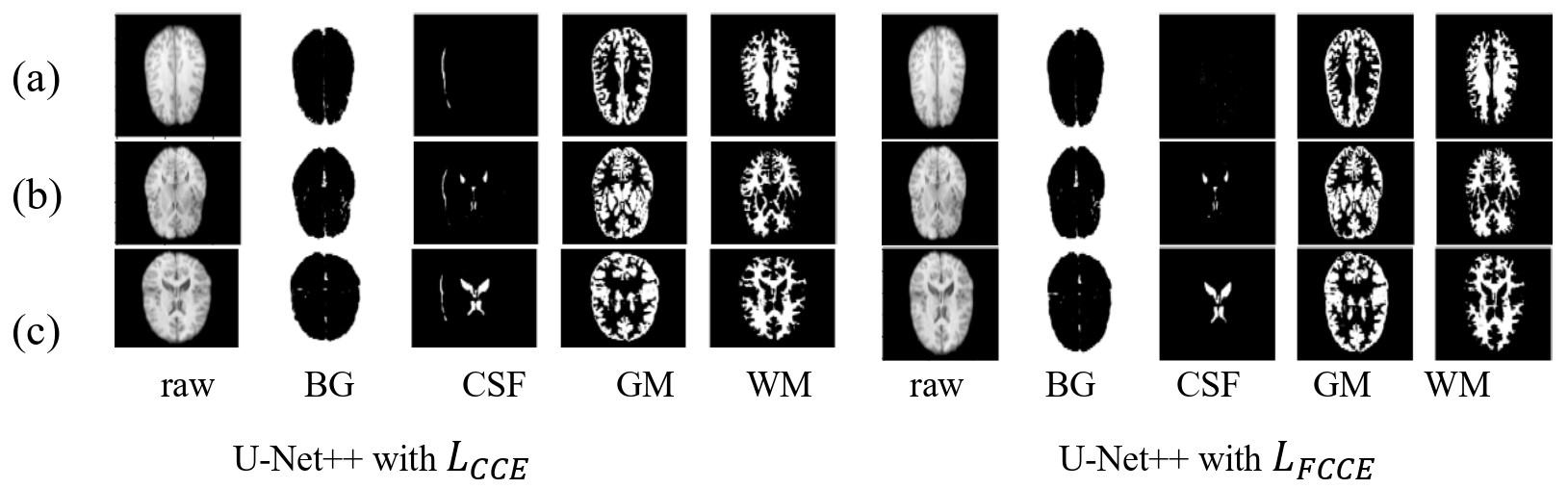}
	\caption{Qualitative segmentation results using U-Net++ with $L_{CCE}$, and U-Net++ with $L_{FCCE}$ (proposed loss function) on the IBSR brain data for the various image (a), (b) and (c). The most left row, comprised the raw image and corresponding predicted image comprises with various tissues such as background (BG), CSF, GM, and WM. The U-Net++ with $L_{FCCE}$ shows the better segmentation in CSF, GM and WM tissues as it handles the uncertainty at the boundary region, as shown in case of CSF tissues.}
	\label{fig:fign6}
\end{figure}

Figure \ref{fig:fign5} and Fig. \ref{fig:fign6}, depicts a qualitative comparison between the results of U-Net and U-Net++ with $L_{CCE}$ and $L_{FCCE}$ loss function for IBSR data respectively. It can be seen that $L_{FCCE}$ with fuzzy logic gives the better results and assigned the image pixels to the desired classes. The objective to obtain the correct segmentation especially for the uncertain case, and comparably some pixels in the image. 

\begin{figure*}  
	\centering
	\begin{subfigure}[t]{0.45\textwidth}
		\centering
		\includegraphics[width=\textwidth]{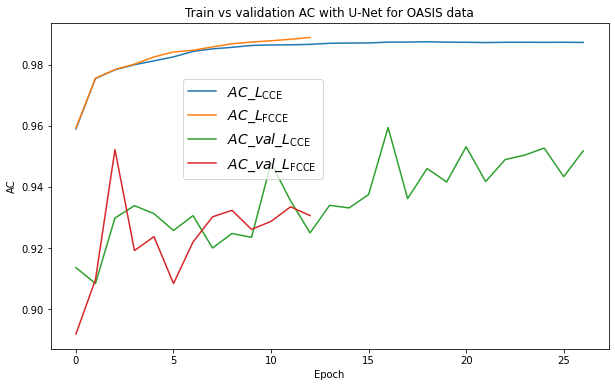}
	\end{subfigure}
	\hfill
	\begin{subfigure}[t]{0.45\textwidth}
		\centering
		\includegraphics[width=\textwidth]{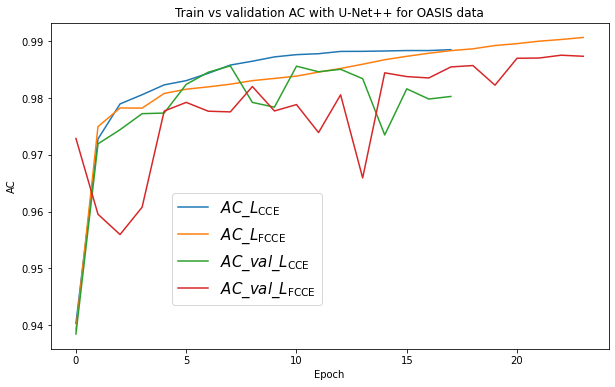}
	\end{subfigure}
	
	\vspace{2em}
	
	\begin{subfigure}[t]{0.45\textwidth}
		\centering
		\includegraphics[width=\textwidth]{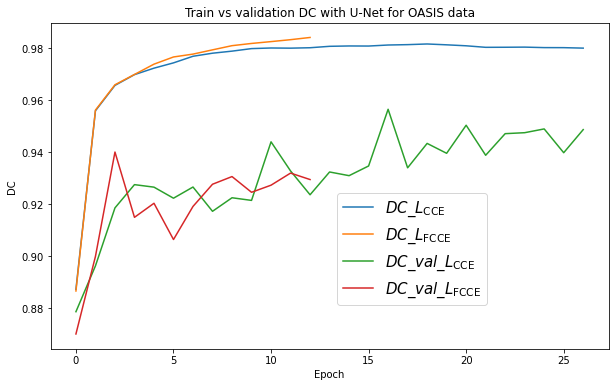}
	\end{subfigure}
	\hfill
	\begin{subfigure}[t]{0.45\textwidth}
		\centering
		\includegraphics[width=\textwidth]{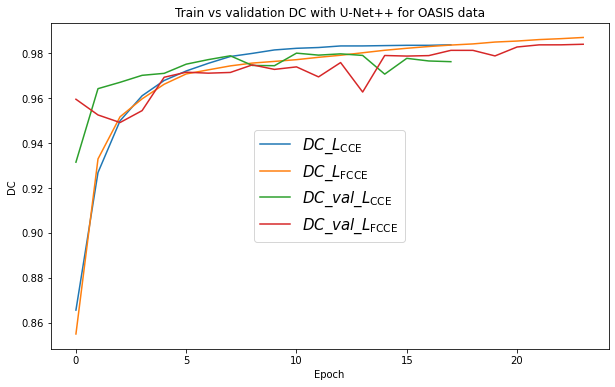}
	\end{subfigure}
	
	\vspace{2em}
	
	\begin{subfigure}[t]{0.45\textwidth}
		\centering
		\includegraphics[width=\textwidth]{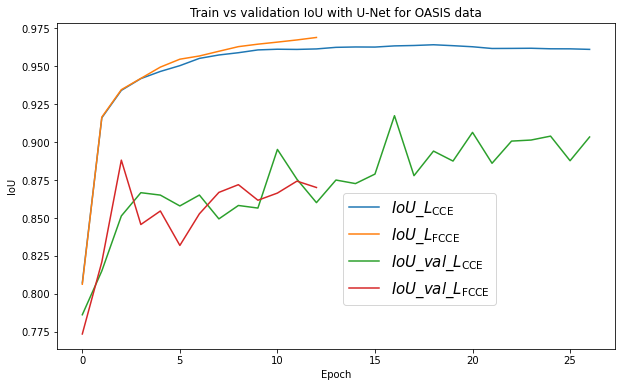}
	\end{subfigure}
	\hfill
	\begin{subfigure}[t]{0.4\textwidth}
		\centering
		\includegraphics[width=\textwidth]{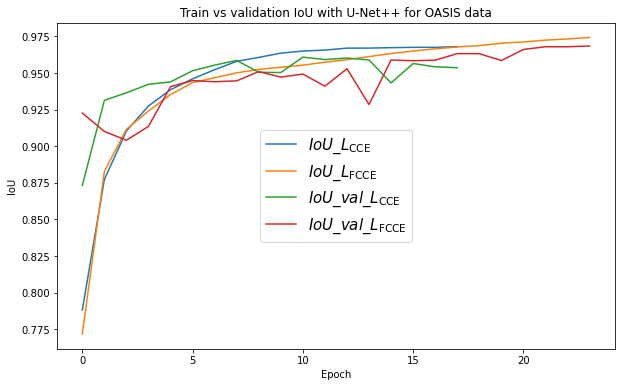}
	\end{subfigure}
	
	\caption{Plot for training versus validation performance for U-Net and U-Net++ architecture for OASIS brain data with $(L_{CCE})$  and $(L_{FCCE})$ loss optimization function along with epoch in term of performance metrices: accuracy, Dice coefficient and Intersection over Union. For OASIS data we used the early stoppage to avoid the overfitting problem. The epoch varies as the model achieve the optimization.  The performance of U-Net and U-Net++ with $(L_{FCCE})$ in term of $AC$, $DC$, $IoU$ performance slightly better with the epoch for OASIS data. During training the model, the U-Net validation with $(L_{FCCE})$ in term of  ${AC}_{{val}}$, ${DC}_{{val}}$, ${IoU}_{{val}}$ providing less or equal results. The U-Net++ validation with $(L_{FCCE})$ in term of ${AC}_{{val}}$, ${DC}_{{val}}$, ${IoU}_{{val}}$ providing more better results in comparison to $(L_{CCE})$. }
	\label{fig:fign7}
\end{figure*}

Further, ablation study for $L_{CCE}$ and $L_{FCCE}$ function presented on OASIS brain dataset. The $20$ cases from the OASIS dataset have been taken and from these $20$ cases, $3520$ brain 2-D images is obtained for the training and testing the model. The corresponding labelled image is used for the validation the model. We test the U-Net and U-Net++ architecture with $L_{CCE}$ and $L_{FCCE}$ for segmentation on all $3520$ images but after some steps its results degrade due to over fitting. To overcome the overfitting we used the EarlyStopping and dropout layers with taking the learning rate 0.0001. Table \ref{tab:tabn2}, comprises the segmentation results and compare the efficacy of segmentation performance in term of a $AC$, $DC$ and $IoU$ and respective validation  ${AC}_{{val}}$, ${DC}_{{val}}$, ${IoU}_{{val}}$. As seen, the U-Net and U-Net++ architecture performed better with $L_{FCCE}$ loss function in comparison to $L_{CCE}$ for $AC$, $DC$ and $IoU$ . The U-Net++ promising the better validation results with fuzzy categorical cross entropy during the model training, however the validation results for U-Net with proposed $L_{FCCE}$  function is not good. For OASIS data, we depicted the training and validation performance during training the model with $L_{CCE}$  and $L_{FCCE}$  optimization function in Fig. \ref{fig:fign7}. 

\begin{table}[H]
	\caption{U-Net and U-Net++ architecture segmentation performance with $L_{CCE}$ and $L_{FCCE}$ loss function for OASIS data in term of $AC$, $DC$, $IoU$ and respective validations ${AC}_{{val}}$, ${DC}_{{val}}$, ${IoU}_{{val}}$. We use EarlyStopping and Batch size 2 for the OASIS data.}
	\centering
	\label{tab:tabn2}
	\begin{tabular}{c c c c}
		\hline
		Algorithm & {Metrics} & {$L_{CCE}$} & {$L_{FCCE}$} \\
		\hline
		\multirow{6}{*}{U-Net} & AC & 0.9873 & 0.9889 \\
		& \( \text{AC}_\text{val} \) & 0.9518 & 0.9306 \\
		& DC & 0.9801 & 0.9842 \\
		& \( \text{DC}_\text{val} \) & 0.9488 & 0.9296 \\
		& IoU & 0.9612 & 0.9691 \\
		& \( \text{IoU}_\text{val} \) & 0.9034 & 0.8701 \\
		\multirow{6}{*}{U-Net++} & AC & 0.9885 & 0.9907 \\
		& \( \text{AC}_\text{val} \) & 0.9803 & 0.9874 \\
		& DC & 0.9837 & 0.9869 \\
		& \( \text{DC}_\text{val} \) & 0.9762 & 0.9839 \\
		& IoU & 0.9680 & 0.9743 \\
		& \( \text{IoU}_\text{val} \) & 0.9536 & 0.9685 \\
		\hline
	\end{tabular}
	
\end{table}

\section{Conclusion }\label{con}

In this work, we have presented a fuzzy categorical cross entropy with integrating the fuzzy entropy in categorical cross-entropy to handle the uncertainty using the fuzzy logic for more accurate image segmentation. With employing fuzzy logic, the proposed optimization function accounts for the inherent uncertainties in pixel classifications and membership degree is computed using the FCM algorithm. The gradient of categorical cross-fuzzy entropy loss functions is computed with combine the gradient of categorical cross-entropy and fuzzy entropy. The proposed loss function has been evaluated on two publicly available brain datasets, IBSR and OASIS, using architectures, U-Net and U-Net++ with training parameter. Experimental results demonstrate consistent performance improvement and trained model with FCCE loss function provided the better results in comparison to CCE optimization function in term of training accuracy, Dice coefficient, and Intersection over Union along with corresponding validation.

The effectiveness of $L_{FCCE}$ in performing comparison to standard $L_{CCE}$ training approaches. By addressing key limitations and exploring future avenues, this work aims to pave the way for developing more reliable in imprecise segmentation models. We believe that these advancements will significantly contribute to image segmentation with uncertainty.



\begin{thebibliography}{99}
	
	\bibitem{Azad2024} Azad, R., Aghdam, E. K., Rauland, A., Jia, Y., Avval, A. H., Bozorgpour, A., ... \& Merhof, D. (2024). Medical image segmentation review: The success of u-net. \textit{IEEE Transactions on Pattern Analysis and Machine Intelligence}.
	
	\bibitem{Azad2023} Azad, R., Heidary, M., Yilmaz, K., Hüttemann, M., Karimijafarbigloo, S., Wu, Y., ... \& Merhof, D. (2023). Loss functions in the era of semantic segmentation: A survey and outlook. \textit{arXiv preprint arXiv:2312.05391}.
	
	\bibitem{Badawy2021} Badawy, S. M., Mohamed, A. E. N. A., Hefnawy, A. A., Zidan, H. E., GadAllah, M. T., \& El-Banby, G. M. (2021). Automatic semantic segmentation of breast tumors in ultrasound images based on combining fuzzy logic and deep learning—A feasibility study. \textit{PloS One}, 16(5), e0251899.
	
	\bibitem{Bezdek1981} Bezdek, J. C. (1981). \textit{Pattern Recognition with Fuzzy Objective Function Algorithms}.
	
	\bibitem{Chaira2015} Chaira, T. (2015). \textit{Medical image processing: Advanced fuzzy set theoretic techniques}. CRC Press.
	
	\bibitem{Chen2022} Chen, Y., Xu, C., Ding, W., Sun, S., Yue, X., \& Fujita, H. (2022). Target-aware U-Net with fuzzy skip connections for refined pancreas segmentation. \textit{Applied Soft Computing}, 131, 109818.
	
	\bibitem{Das2020} Das, R., Sen, S., \& Maulik, U. (2020). A survey on fuzzy deep neural networks. \textit{ACM Computing Surveys (CSUR)}, 53(3), 1-25.
	
	\bibitem{Fawzi2021} Fawzi, A., Achuthan, A., \& Belaton, B. (2021). Brain image segmentation in recent years: A narrative review. \textit{Brain Sciences}, 11(8), 1055.
	
	\bibitem{Gupta2023} Gupta, A., Verma, H., Prasad, M., Kirar, J. S., \& Lin, C. T. (Eds.). (2023). \textit{Computational Intelligence Aided Systems for Healthcare Domain}. CRC Press.
	
	\bibitem{Hu2021} Hu, M., Zhong, Y., Xie, S., Lv, H., \& Lv, Z. (2021). Fuzzy system based medical image processing for brain disease prediction. \textit{Frontiers in Neuroscience}, 15, 714318.
	
	\bibitem{Huang2022} Huang, K., Zhang, Y., Cheng, H. D., \& Xing, P. (2022, December). Trustworthy breast ultrasound image semantic segmentation based on fuzzy uncertainty reduction. \textit{Healthcare}, 10(12), 2480.
	
	\bibitem{Huang2021} Huang, K., Zhang, Y., Cheng, H. D., Xing, P., \& Zhang, B. (2021). Semantic segmentation of breast ultrasound image with fuzzy deep learning network and breast anatomy constraints. \textit{Neurocomputing}, 450, 319-335.
	
	\bibitem{Jadon2020} Jadon, S. (2020, October). A survey of loss functions for semantic segmentation. In \textit{2020 IEEE Conference on Computational Intelligence in Bioinformatics and Computational Biology (CIBCB)} (pp. 1-7). IEEE.
	
	\bibitem{Janocha2017} Janocha, K., \& Czarnecki, W. M. (2017). On loss functions for deep neural networks in classification. \textit{arXiv preprint arXiv:1702.05659}.
	
	\bibitem{Karakose2024} Karaköse, E. (2024). An Efficient Satellite Images Classification Approach Based on Fuzzy Cognitive Map Integration with Deep Learning Models Using Improved Loss Function. \textit{IEEE Access}.
	
	\bibitem{Landgraf2023} Landgraf, S., Hillemann, M., Wursthorn, K., \& Ulrich, M. (2023). U-CE: Uncertainty-aware cross-entropy for semantic segmentation. \textit{arXiv preprint arXiv:2307.09947}.
	
	\bibitem{Liu2020} Liu, L., Cheng, J., Quan, Q., Wu, F. X., Wang, Y. P., \& Wang, J. (2020). A survey on U-shaped networks in medical image segmentations. \textit{Neurocomputing}, 409, 244-258.
	
	\bibitem{Price2019} Price, S. R., Price, S. R., \& Anderson, D. T. (2019, June). Introducing fuzzy layers for deep learning. In \textit{2019 IEEE International Conference on Fuzzy Systems (FUZZ-IEEE)} (pp. 1-6). IEEE.
	
	\bibitem{Punn2022} Punn, N. S., \& Agarwal, S. (2022). Modality specific U-Net variants for biomedical image segmentation: a survey. \textit{Artificial Intelligence Review}, 55(7), 5845-5889.
	
	\bibitem{Ronneberger2015} Ronneberger, O., Fischer, P., \& Brox, T. (2015). U-net: Convolutional networks for biomedical image segmentation. In \textit{Medical Image Computing and Computer-Assisted Intervention–MICCAI 2015: 18th International Conference} (pp. 234-241). Springer International Publishing.
	
	\bibitem{Shannon1948} Shannon, C. E. (1948). A mathematical theory of communication. \textit{The Bell System Technical Journal}, 27(3), 379-423.
	
	\bibitem{Sharma2019} Sharma, T., Singh, V., Sudhakaran, S., \& Verma, N. K. (2019, June). Fuzzy based pooling in convolutional neural network for image classification. In \textit{2019 IEEE International Conference on Fuzzy Systems (FUZZ-IEEE)} (pp. 1-6). IEEE.
	
	\bibitem{Subhashini2022} Subhashini, L. D. C. S., Li, Y., Zhang, J., \& Atukorale, A. S. (2022). Integration of fuzzy logic and a convolutional neural network in three-way decision-making. \textit{Expert Systems with Applications}, 202, 117103.
	
	\bibitem{Verma2014} Verma, H., Agrawal, R. K., \& Kumar, N. (2014). Improved fuzzy entropy clustering algorithm for MRI brain image segmentation. \textit{International Journal of Imaging Systems and Technology}, 24(4), 277-283.
	
	\bibitem{Verma2016} Verma, H., Agrawal, R. K., \& Sharan, A. (2016). An improved intuitionistic fuzzy c-means clustering algorithm incorporating local information for brain image segmentation. \textit{Applied Soft Computing}, 46, 543-557.
	
	\bibitem{Verma2023} Verma, H., Gupta, A., Kirar, J. S., Prasad, M., \& Lin, C. T. (2023). Introduction to computational methods: Machine and deep learning perspective. In \textit{Computational Intelligence Aided Systems for Healthcare Domain} (pp. 1-32). CRC Press.
	
	\bibitem{Verma2020} Verma, O. P., \& Hooda, H. (2020). A novel intuitionistic fuzzy co-clustering algorithm for brain images. \textit{Multimedia Tools and Applications}, 79(41), 31517-31540.
	
	\bibitem{Wang2022} Wang, R., Lei, T., Cui, R., Zhang, B., Meng, H., \& Nandi, A. K. (2022). Medical image segmentation using deep learning: A survey. \textit{IET Image Processing}, 16(5), 1243-1267.
	
	\bibitem{Yang2021} Yang, Z., Xu, P., Yang, Y., \& Kang, B. (2021). Noise robust intuitionistic fuzzy c‐means clustering algorithm incorporating local information. \textit{IET Image Processing}, 15(3), 805-817.
	
	\bibitem{Zadeh1965} Zadeh, L. A. (1965). Fuzzy sets. \textit{Information and Control}, 8(3), 338-353.
	
	\bibitem{Zadeh1996} Zadeh, L. A., Klir, G. J., \& Yuan, B. (1996). \textit{Fuzzy sets, fuzzy logic, and fuzzy systems: selected papers (Vol. 6)}. World Scientific.
	
	\bibitem{Zhou2018} Zhou, Z., Rahman Siddiquee, M. M., Tajbakhsh, N., \& Liang, J. (2018). Unet++: A nested u-net architecture for medical image segmentation. In \textit{Deep Learning in Medical Image Analysis and Multimodal Learning for Clinical Decision Support: 4th International Workshop, DLMIA 2018, and 8th International Workshop, ML-CDS 2018} (pp. 3-11). Springer International Publishing.
	
	\bibitem{Zhou2019} Zhou, Z., Siddiquee, M. M. R., Tajbakhsh, N., \& Liang, J. (2019). Unet++: Redesigning skip connections to exploit multiscale features in image segmentation. \textit{IEEE Transactions on Medical Imaging}, 39(6), 1856-1867.
	
\end{thebibliography}
\end{document}